\pdfoutput=1
\documentclass[11pt]{article}

\usepackage[utf8]{inputenc}
\usepackage[T1]{fontenc}
\usepackage{lmodern}
\usepackage{libertine}
\usepackage{microtype}

\usepackage[left=1in,right=1in,top=1in,bottom=1in]{geometry}
\usepackage{parskip}

\usepackage{amsmath,amsthm,amstext,amsfonts,amssymb,mathrsfs,mathtools,bbm}
\usepackage{nicefrac}

\usepackage{graphicx}
\usepackage{caption}
\usepackage{subcaption}
\usepackage{float}
\usepackage{wrapfig}
\graphicspath{{figs/}}

\usepackage{booktabs}
\usepackage{array}
\usepackage{multirow}
\usepackage{tabularx}

\usepackage[table]{xcolor}
\usepackage[colorlinks=true,citecolor=teal,linkcolor=teal,urlcolor=teal]{hyperref}
\usepackage{url}

\usepackage[ruled,vlined]{algorithm2e}
\usepackage[noend]{algpseudocode}

\usepackage{enumitem}
\usepackage{changepage}
\usepackage{ragged2e}
\usepackage{dsfont}
\usepackage{tikz}
\usetikzlibrary{arrows,shapes,chains,matrix,positioning,scopes,patterns,calc}

\usepackage[numbers,sort&compress]{natbib}
\usepackage[capitalize]{cleveref}

\newcommand{\dk}[1]{}

\title{GSpaRC: Gaussian Splatting for\\Real-time Reconstruction of RF Channels}

\author{
Bhavya Sai Nukapotula$^{1}$ \quad Rishabh Tripathi$^{1}$ \quad Seth Pregler$^{1}$ \\[0.2em]
Dileep Kalathil$^{1}$ \quad Srinivas Shakkottai$^{1}$ \quad Tedd Rappaport$^{2}$ \\[0.4em]
{\small $^{1}$Texas A\&M University \quad $^{2}$New York University}
}

\begin{document}

\date{}
\maketitle

\begin{abstract}
Channel state information (CSI) is essential for adaptive beamforming and maintaining robust links in wireless communication systems. However, acquiring CSI incurs significant overhead, consuming up to 25\% of spectrum resources in 5G networks due to frequent pilot transmissions at millisecond-scale intervals. Recent approaches aim to reduce this burden by reconstructing CSI from spatiotemporal RF measurements, such as signal strength and  direction-of-arrival. While effective in offline settings, these methods often suffer from inference latencies in the 5--100~ms range, making them impractical for real-time systems.  We present GSpaRC: Gaussian Splatting for Real-time Reconstruction of RF Channels, a method that achieves accurate channel reconstruction with latency in the low-millisecond regime or below.  GSpaRC represents the RF environment using a compact set of 3D Gaussian primitives, each parameterized by a lightweight neural model augmented with physics-informed features such as distance-based attenuation. Unlike traditional vision-based splatting pipelines, GSpaRC is tailored for RF reception: it employs an equirectangular projection onto a hemispherical surface centered at the receiver to reflect omnidirectional antenna behavior. A custom CUDA pipeline enables fully parallelized directional sorting, splatting, and rendering across frequency and spatial dimensions.  Evaluated on multiple RF datasets, GSpaRC achieves similar CSI reconstruction fidelity to recent state-of-the-art methods while reducing training and inference time by over an order of magnitude. These results illustrate that modest GPU computation can substantially reduce pilot overhead, making GSpaRC a scalable low-latency approach for channel estimation in 5G and future wireless systems. The project website is available here: \href{https://nbhavyasai.github.io/GSpaRC/}{GSpaRC}.


\end{abstract}

\section{Introduction}


Machine learning is increasingly moving into the real-time wireless stack, where computation must support channel estimation and related physical-layer decisions under tight latency budgets. This trend is reinforced by GPU-accelerated RAN platforms such as NVIDIA Aerial, which make it feasible to place learned components inside these PHY-layer loops while meeting latency constraints on the order of a few milliseconds, corresponding to transmission time intervals (TTIs)~\cite{nvidia_aerial}. One especially promising opportunity is channel modeling and prediction. Wireless propagation is fundamentally geometric: the channel depends on the layout of surfaces, blockers, and antennas, as well as on how signals interact with them through reflection, diffraction, scattering, and shadowing. This geometric view has long motivated ray-tracing and ray-launching methods for wireless propagation modeling~\cite{seidel1992ray,durgin1997advanced,nguyen2014evaluation}. A useful learned representation should therefore do more than interpolate past measurements or extrapolate a short time series. It should be geometry-aware, reusable across receiver locations, and informative for downstream wireless tasks.


\begin{figure}[t]
    \centering
    \includegraphics[width=0.75\textwidth]{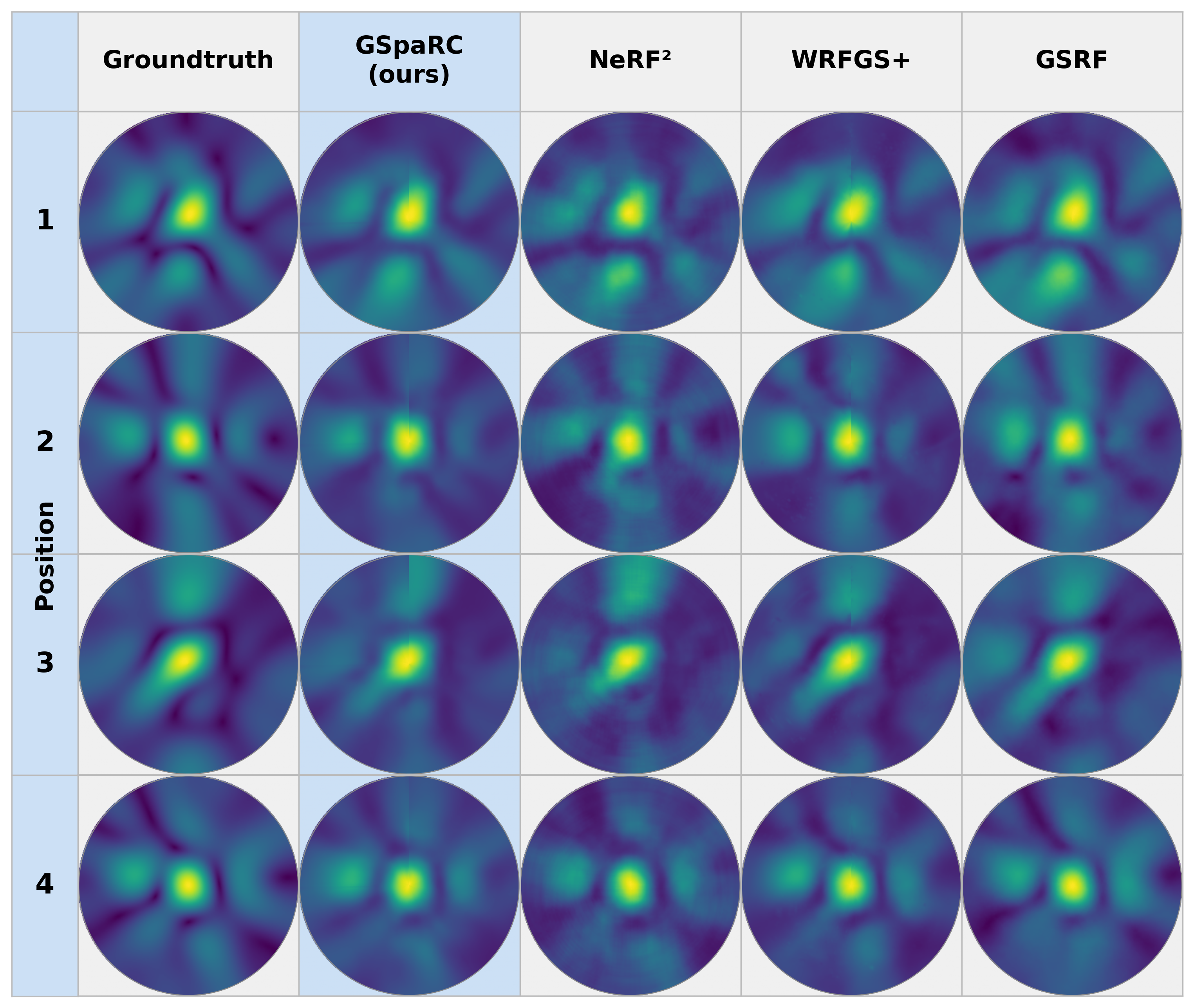}
    \caption{{Representative hemispherical spectrum reconstructions for four transmitter positions on the RFID dataset. All methods produce visually similar results; GSpaRC (highlighted column) achieves this at low-millisecond latency.}}
    \label{fig:comparison}
\end{figure}

This viewpoint is important because channel acquisition consumes substantial signaling resources, often on the order of \(10\%-25\%\) of time-frequency resources in current systems. Different pilots also serve different roles. Demodulation reference signals (DMRS) support fine time and frequency estimation for symbol detection, while sounding reference signals (SRS), and more generally larger-scale sounding procedures, collect measurements over different locations and times to support link adaptation, modulation and coding scheme (MCS) selection, beam management, and physical resource block (PRB) allocation. This motivates a geometrically aligned RF representation that can account for measurements taken at different locations and rapidly predict channel structure when geometry changes slowly relative to fine-scale fading. Such a representation can reduce pilot overhead and improve spectral efficiency.

Recent NeRF- and Gaussian-based RF methods can already produce plausible hemispherical reconstructions \cite{zhao2023nerf2,lu2024newrf,wen2025neuralrepresentationwirelessradiation,yang2025scalable3dgaussiansplattingbased,yang2025gsrf}. Indeed, Fig.~\ref{fig:comparison} shows representative examples in which several methods (including our own) look qualitatively similar on a real-world RFID data set~\cite{zhao2023nerf2}. The main challenge is therefore not only fidelity, but operating point, in that the representation must be geometrically grounded, fast enough  for real-time inference (timescale of a few TTIs), and useful for downstream control. It must also provide a confidence estimate, since a prediction is only useful if we know when it can be trusted.

3D Gaussian splatting~\cite{kerbl20233d} is attractive in this setting because it provides an explicit 3D representation that can capture persistent shadowing, diffraction, and multipath structure rather than only short-term signal correlations. This inductive bias is especially natural for downlink prediction, where a fixed transmitter illuminates the scene and the receiver moves through space. In that regime, the learned field can encode stable geometric effects and provide location-dependent confidence estimates that are important for downstream tasks such as MCS selection and resource allocation, which are the primary focus of our dataset generation and downstream evaluations.

In this paper, we present \emph{GSpaRC}, a real-time Gaussian-splatting framework for RF reconstruction and channel prediction. Our contributions are fourfold. \textit{First,} we develop a geometrically aware RF representation based on anisotropic 3D Gaussians with small MLP decoders that map the learned field to either directional spectra or complex channel estimates. \textit{Second,} we introduce an RF-specific rendering pipeline built around a receiver-centered equirectangular hemispherical projection and a custom CUDA implementation, enabling low-millisecond inference (around 300 $\mu$s to 5 ms on a modest GPU, depending on scene complexity). \textit{Third,} because the representation is spatially grounded, it naturally supports calibrated confidence estimates at each query location.  We also integrate the full pipeline with the open-source NVIDIA Sionna platform~\cite{sionna} and evaluate it across a scene specified by point-cloud geometry and Sionna-generated RF data. \textit{Fourth,} on the real-world Argos dataset~\cite{shepard2016understanding}, we show that this geometrically grounded representation is useful beyond reconstruction in that over \(70\%\) of reconstructed points are assigned high confidence, and the confidence score identifies which predicted points are reliable enough for downstream tasks such as MCS selection and channel equalization for accurate symbol detection.

\begin{table}[h]
\centering
\small
\caption{Representative comparison on a synthetic Sionna-generated conference room data set. GSpaRC achieves higher accuracy, faster rendering, and provides a per-position confidence score.}
\begin{tabular}{@{}lccccc@{}}
\toprule
\textbf{Algorithm} & \shortstack{\textbf{Train} \\ \textbf{(hrs)} $\downarrow$} & \shortstack{\textbf{Render} \\ \textbf{(ms)} $\downarrow$} & \textbf{SSIM} $\uparrow$ & \shortstack{\textbf{PSNR} \\ \textbf{(dB)} $\uparrow$} & \shortstack{\textbf{Confidence} \\ \textbf{Score}} \\
\midrule
GSRF   & 1.22 & 23   & 0.70          & 25.76          & $\times$ \\
\midrule
\textbf{GSpaRC} & \textbf{1.17} & \textbf{3.8} & \textbf{0.78} & \textbf{27.16} & \textbf{\checkmark} \\
\bottomrule
\end{tabular}
\label{tab:time_comparison_intro}
\end{table}
Table~\ref{tab:time_comparison_intro} presents a representative operating point achieved by GSpaRC on a synthetic data set that we generated using Sionna for a conference room scene with many occluding objects. GSpaRC outperforms GSRF in both SSIM and PSNR while rendering over $6\times$ faster.  These results are obtained using a modest NVIDIA RTX 2080 Ti GPU (circa 2018).  Critically, GSpaRC is the only method that produces a per-position confidence score, allowing it to identify which predicted positions can replace conventional pilots.






\begin{figure}[tb]
\centering
\includegraphics[width=0.75\textwidth]{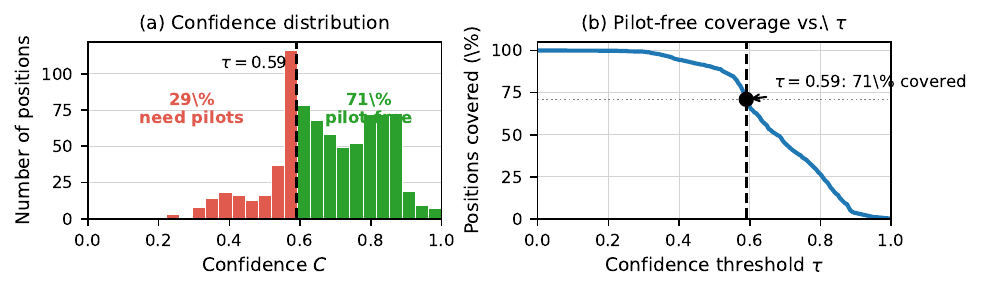}
\caption{Example confidence distribution and pilot-free coverage on the Argos dataset (800 test positions). (a) Positions with $C \ge \tau{=}0.59$ (green) predicted by GSpaRC require no pilot transmission; only 30\% fall back to pilots. (b) Coverage curve showing pilot-free fraction vs.\ confidence threshold $\tau$.}
\label{fig:confidence_coverage}
\end{figure}
We illustrate the importance of confidence estimation using the real-world Argos dataset. As shown in Figure~\ref{fig:confidence_coverage}, confidence reflects how reliably the learned Gaussian field can reconstruct the channel at a given location: positions where the propagation geometry is well-represented receive high confidence, while positions in difficult multipath or shadowed regions receive low confidence (see Sec.~\ref{sec:gs_wireless} for the formal definition and normalization). Empirically, a confidence threshold of 0.59 provides sufficient accuracy for downstream tasks such as MCS selection and channel equalization. Our approach identifies the \(\approx 70\%\) of receiver positions that exceed this threshold, i.e., pilot transmissions can be avoided for most locations.


\section{Related Work}

Recent efforts have extended Neural Radiance Fields (NeRF) \cite{mildenhall2021nerf} to model radio propagation in complex 3D environments. \textbf{NeRF\textsuperscript{2}} \cite{zhao2023nerf2} is a notable example that uses a NeRF-like architecture with two neural networks. It takes as input a virtual transmitter position, the direction from receiver to transmitter, and the transmitter location. The network predicts signal strength and attenuation, incorporating attenuation into the rendering equation as opacity. This adapts volumetric rendering from computer graphics to RF propagation. While NeRF\textsuperscript{2} generalizes well from sparse measurements, it requires several hours of training and has high latency, often hundreds of milliseconds per frame, which limits real-time use.

\textbf{NeWRF}~\cite{lu2024newrf} extends NeRF for wireless channel prediction by adding a $1/d$ path loss term and using direction-of-arrival information. It uses ray casting guided by DoA estimates and a ray search procedure to generalize to unseen receiver locations. While effective in fixed-transmitter, mobile-receiver settings, it also requires hours of training and has inference latency in the hundreds of milliseconds, making it too slow for real-time use.

\textbf{WRFGS}~\cite{wen2025neuralrepresentationwirelessradiation} applies Gaussian splatting to RF modeling by representing the radio field with spatial Gaussians and using a neural network to predict signal values and attenuation. WRFGS+ improves accuracy with a more expressive network that also predicts opacity, scaling, and rotation. This added complexity increases computation, requiring hours of training and about 8\,ms inference time, which reduces the efficiency benefits of Gaussian splatting.

\textbf{GSRF}~\cite{yang2025gsrf} is the strongest prior Gaussian-splatting baseline that we compare against. It builds on the same authors' earlier RFSPM formulation~\cite{yang2025scalable3dgaussiansplattingbased}, which served as a prototype for GSRF.   RFSPM used per-Gaussian MLPs to model directional emission, while  GSRF improves on that design by replacing the per-Gaussian MLPs with a Fourier--Legendre basis functions per Gaussian, and by adding a frequency-domain loss during training. These changes improve reconstruction quality, but rendering still requires evaluating the basis functions for each Gaussian and computing eigenvalue decompositions for Gaussian radii. As a result, the method remains computationally heavy, with about 23\,ms latency.

\section{Preliminaries and Problem Formulation}

\subsection{Multipath Channel Representation}
\label{subsec:multipath}
In a static narrowband propagation environment, the physical channel is characterized by a finite set of
multipath coefficients $\{a_i,\tau_i\}$, where $a_i$ and $\tau_i$ denote the complex gain and delay of the
$i$th path.  These parameters uniquely determine the baseband-equivalent impulse response
\[
    h_b(\tau)=\sum_i a_i e^{-j2\pi f_c \tau_i}\,\delta(\tau-\tau_i),
\]
which fully specifies how the transmitted waveform is filtered by the propagation medium.  
The corresponding continuous-time input--output relation is the LTI convolution
$y_b(t)= (h_b * x_b)(t)$.

Sampling this model at the symbol rate yields a discrete-time convolution with tap values $\{h[\ell]\}$ that
capture the resolvable multipath delays, along with noise samples $w[m]$ obtained from filtered AWGN.
In a SIMO system with $N_r$ antennas, each element observes the same symbol sequence but through a
different discrete-time impulse response.  Under the narrowband (frequency-flat) assumption appropriate
for a single OFDM subcarrier, each impulse response collapses to a single complex gain $h_r$, so the
per-antenna model is
\[
    y_r[m] = h_r\,x[m] + w_r[m], \qquad r=1,\dots,N_r,
\]
where $h_r$ is the per-antenna channel coefficient formed by coherent combination of the
multipath components at that element~\cite{Tse_Viswanath_2005}.

\subsection{3D Gaussian Splatting}
\label{subsec:3dgs}
Gaussian Splatting (GS) replaces the implicit volumetric MLP of NeRF with an explicit set of anisotropic 3D Gaussian primitives $(\mu_k, \Sigma_k, \alpha_k, c_k)$. Each primitive is defined by: (i) a mean $\mu_k\!\in\!\mathbb{R}^3$ giving the 3D center of the Gaussian; (ii) a covariance $\Sigma_k\!\in\!\mathbb{R}^{3\times 3}$ encoding its spatial extent, orientation, and anisotropy; (iii) an opacity $\alpha_k\!\in\![0,1]$ determining the accumulation weight along the viewing ray; and (iv) a color $c_k\!\in\!\mathbb{R}^3$ specifying the RGB radiance emitted by the primitive. The density field contributed by the $k$-th Gaussian is {\small $G_k(x)=\exp[-\tfrac{1}{2}(x-\mu_k)^\top\allowbreak \Sigma_k^{-1}(x-\mu_k)]$}. To ensure that $\Sigma_k$ remains positive semi-definite and numerically stable during optimization, it is parameterized as {\small $\Sigma_k = R_k S_k S_k^\top R_k^\top$}, where $R_k\!\in\!SO(3)$ is a rotation matrix controlling orientation and $S_k$ is a diagonal scale matrix controlling the principal-axis radii.

Under a camera projection $\Pi:\mathbb{R}^3\!\rightarrow\!\mathbb{R}^2$, each 3D Gaussian induces a 2D elliptical footprint on the image plane. This footprint is obtained by locally linearizing $\Pi$ around the Gaussian mean and propagating the 3D covariance through the projection. The resulting 2D covariance is {\small $\tilde{\Sigma}_k \propto J_{\Pi}(\mu_k)\,\Sigma_k\,J_{\Pi}(\mu_k)^\top$}, where $J_{\Pi}(\mu_k)$ is the Jacobian of the affine approximation of the projective transformation. The corresponding normalized image-plane kernel is {\small $\mathcal{G}_k(u)=\exp[-\tfrac{1}{2}(u-\Pi(\mu_k))^\top\allowbreak \tilde{\Sigma}_k^{-1}(u-\Pi(\mu_k))]$}, which determines how strongly the Gaussian contributes to pixel $u$.

Rendering is performed by depth-sorting the projected Gaussians and applying visibility-aware front-to-back $\alpha$-compositing. The color at pixel $u$ is {\small $\hat{C}(u)=\sum_k [T_k\,\alpha_k\,\mathcal{G}_k(u)\,c_k]$}, where the transmittance is {\small $T_k=\prod_{j<k}(1-\alpha_j\,\mathcal{G}_j(u))$}. This explicit, differentiable representation provides high-fidelity novel-view synthesis with real-time rendering performance and significantly reduced computational cost compared to volumetric ray marching.

\subsection{Problem Formulation}
\label{subsec:problem}

\begingroup\sloppy
Our objective is to develop a physically grounded representation of the RF propagation environment that can synthesize accurate channel characteristics at arbitrary 3D query locations with sub-millisecond latency. As described in Sec.~\ref{subsec:multipath}, the wireless channel is formed by the coherent superposition of a finite set of multipath components. Their complex amplitudes and delays encode geometry-dependent phase evolution, making the channel extremely sensitive to small variations in the scene or carrier frequency. Modeling this joint multipath--geometry--directionality structure is challenging because RF observations are sparse, and signal strength exhibits a strict distance-dependent attenuation (approximately $1/d^2$), a behavior not present in vision-based measurements. Reliable modeling further requires directional responses that are often low-resolution, noisy, or only partially observed.

From a computational standpoint, synthesizing channel characteristics is demanding. It requires repeated steering operations, coherent phase aggregation, and inversion of directional projections. These steps are numerically sensitive and, when implemented naively, exceed the strict timing budget of PHY-layer control. Classical Gaussian Splatting cannot be used directly because RF has no camera intrinsics, no projective geometry, and requires custom CUDA kernels to propagate complex-valued fields, enforce phase consistency, and compute spectrum-to-channel mappings. A practical system must also provide a statistically meaningful confidence metric to decide when synthesized channels may safely replace pilot-based estimates. Existing RF modeling, reconstruction, and localization techniques cannot produce continuous 3D channel fields, do not preserve the complex-valued structure required for coherent communication, and miss real-time latency targets by one to two orders of magnitude. These limitations motivate the central question: \emph{Can we construct an RF-aware Gaussian Splatting framework that captures geometry-induced multipath propagation, produces physically consistent channel characteristics, and operates with a low-millisecond latency for integration into real communication systems?}
\endgroup

\section{Gaussian Splatting for Wireless Scene Reconstruction}
\label{sec:gs_wireless}

\begin{figure}[h]
\centering
\includegraphics[width=\textwidth]{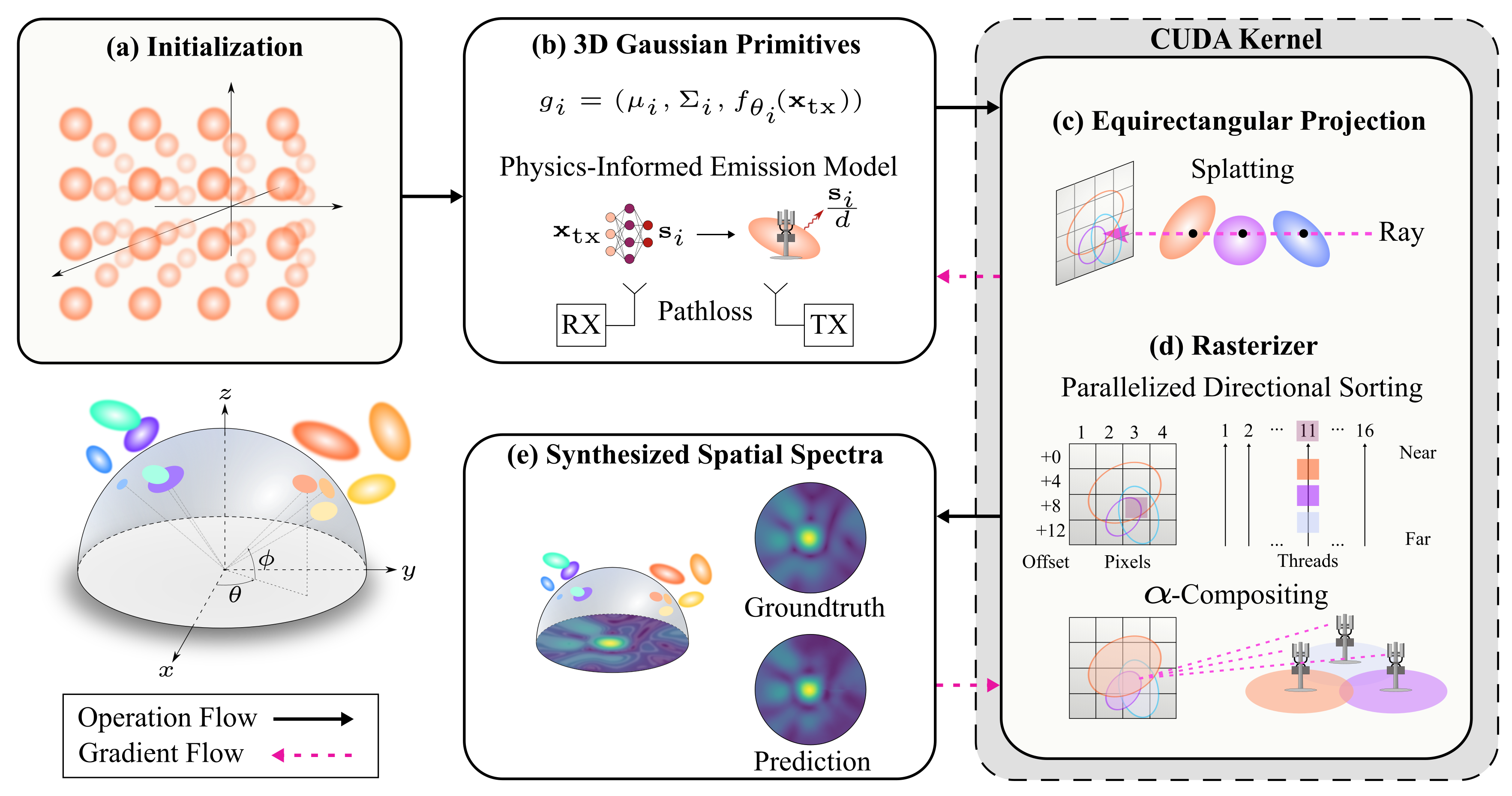}
\caption{Overview of the GSpaRC pipeline. Anisotropic 3D Gaussians are splatted at the receiver position; each Gaussian's complex emission is predicted by a lightweight MLP. In the fixed-TX setting evaluated here, the transmitter position $\mathbf{x}_{\mathrm{tx}}$ is constant, so the MLP is conditioned on the receiver position $\mathbf{x}_{\mathrm{rx}}$ (replacing $\mathbf{x}_{\mathrm{tx}}$ in the figure). The rasterizer accumulates contributions into a directional spectrum, supervised against ground-truth measurements during training.}
\label{fig:pipeline}
\end{figure}

Building on the channel model in Sec.~\ref{subsec:multipath}, we introduce
\emph{GSpaRC}, a Gaussian-splatting-based framework tailored for RF scene reconstruction. GSpaRC
constructs a differentiable representation of the propagation environment and provides a rendering
pipeline that maps this representation to directional spectra and, subsequently, to channel
characteristics (Fig.~\ref{fig:pipeline}). The design is engineered to reflect RF propagation physics while meeting the stringent
latency requirements of real-time communication systems.

\subsection{Scene Representation}

\paragraph*{Gaussian Parameterization.}
GSpaRC models the propagation environment using anisotropic 3D Gaussian primitives that collectively 
form a continuous and differentiable field. Each primitive $i$ is parameterized by 
$(\mu_i, \Sigma_i, \sigma_i, s_i)$: the center $\mu_i \in \mathbb{R}^3$ specifies spatial location, 
$\Sigma_i$ is the covariance matrix formulated as in Sec.~\ref{subsec:3dgs}, 
$\sigma_i$ denotes learned opacity, and $s_i \in \mathbb{C}$ encodes its complex-valued radiative 
contribution in the rendered directional spectrum.

\paragraph*{Directional Neural Predictor.}
Each Gaussian is equipped with a lightweight neural predictor for directional propagation.
A compact MLP $f_\vartheta\colon\mathbb{R}^3\!\to\!\mathbb{R}^2$ takes the receiver position
$\mathbf{x}_{\mathrm{rx}}$ as input. The transmitter contributes a log-distance path-loss
bias: letting $d_i = \|\mathbf{x}_{\mathrm{tx}} - \boldsymbol{\mu}_i\|$ be the distance from the
transmitter to Gaussian $i$, the complex emission is
\begin{equation}
    s_i = \sigma\!\left(f_\vartheta(\mathbf{x}_{\mathrm{rx}}) - \log d_i\right),
    \label{eq:emission}
\end{equation}
where $\sigma$ denotes the sigmoid function applied element-wise to the real and imaginary
parts. This formulation decouples location-dependent receiver sensitivity (learned by the MLP)
from geometric path-loss (encoded analytically via $\log d_i$), providing a learned
complex-valued directional response that forms the intermediate field used to render
directional spectra and channel characteristics.

\paragraph*{Initialization.}
The Gaussian field is initialized in a geometry-agnostic manner to ensure stable optimization. Primitive 
centers $\mu_i$ are uniformly distributed over the spatial domain, and covariances $\Sigma_i$ are set to 
isotropic values following standard 3D Gaussian Splatting practice. Rotations are initialized to identity, 
opacities $\sigma_i$ are set to small constants to avoid early dominance, and MLP parameters are sampled 
from standard normal distributions. This yields a numerically stable baseline from which the representation 
can adapt to multipath structure and directional propagation behavior.

\subsection{Rendering and Signal Formation}
\paragraph*{Rendered Spectrum.}
GSpaRC renders a complex-valued directional spectrum $\hat{\mathbf{z}} \in \mathbb{C}^{H \times W}$ on an equirectangular hemispherical grid, where each pixel encodes the aggregate complex contribution from all Gaussians in that look direction. During training, $\hat{\mathbf{z}}$ is directly supervised against the ground-truth spectrum image using a combination of L1 and SSIM losses. For downstream CSI reconstruction, the complex channel estimate is obtained as 
\begin{equation}
    \hat{h} = \frac{1}{HW}\sum_{u,v} \hat{\mathbf{z}}[u,v],
    \label{eq:spatial_mean}
\end{equation}
which aggregates directional contributions into a complex scalar, compatible with supervised training against CSI ground truth.

\paragraph*{Directional Projection and Splatting.}
For receiver pose $\mathbf{x}_{\mathrm{rx},i}$, primitives are transformed into receiver-centric 
coordinates using
\[
    \mathbf{x}_i' = V(\mu_i - \mathbf{x}_{\mathrm{rx},i}),
\]
with $V \in \mathbb{R}^{3\times 3}$ aligning the receiver at the origin and its viewing axis with $z$.
Directional sampling is performed on a hemispherical grid with $1^\circ$ resolution. Each direction is 
mapped to an equirectangular image coordinate via
\vspace{-2pt}
\[
\begin{bmatrix}
p_x \\ p_y
\end{bmatrix}
=
\begin{bmatrix}
\left(\frac{\arctan2(x_i', z_i')}{\pi}+1\right)\frac{W}{2} \\[4pt]
2\,\arcsin\!\left(\frac{y_i'}{\sqrt{x_i'^2+y_i'^2+z_i'^2}}\right)\frac{H}{\pi}
\end{bmatrix}.
\]
\vspace{-2pt}
Primitives are rasterized onto this directional image through a custom CUDA kernel. Their 2D elliptical 
footprints are obtained by propagating $\Sigma_i$ through the projection Jacobian:
\[
    \Sigma_i' = J\, V\, \Sigma_i\, V^\top J^\top,
\]
after which the depth dimension is discarded.

\paragraph*{Differentiable Rendering.}
For each direction $(\theta,\phi)$, GSpaRC aggregates contributions from all intersecting Gaussians. 
After sorting by distance to the receiver, the MLP $f_\vartheta$ provides a complex coefficient $s_i$ 
for each primitive. The rendered signal is computed via an RF-adapted volumetric compositing rule:
\begin{equation}
    S(\theta,\phi)
    =
    \sum_{i \in \mathcal{I}(\theta,\phi)}
    \left[
        \alpha_i
        \prod_{j < i} (1 - \alpha_j)
        \frac{s_i}{d_i}
    \right],
\end{equation}
where $\mathcal{I}(\theta,\phi)$ is the depth-ordered set of intersecting primitives, 
$\alpha_i = \sigma_i G_i'$ is the projected opacity, and 
$d_i = \|\mu_i - \mathbf{x}_{\mathrm{tx}}\|_2$ imposes a free-space attenuation factor. 
This formulation extends alpha compositing to complex-valued RF propagation and remains fully 
differentiable, enabling gradient flow through geometric, radiometric, and neural components of the 
pipeline. The impact of the distance-based attenuation term $1/d_i$ is evaluated in Sec.~\ref{subsec:ablation-distance}.

\clearpage
\begin{algorithm}[H]
\small
\caption{GSpaRC: Rendering-Based Spectrum Training}
\label{alg:gsparc_spectrum}
\LinesNotNumbered
\SetKwInOut{Input}{Input}
\SetKwInOut{Output}{Output}

\Input{$T$: Number of training steps; $D$: Dataset of $(\mathbf{z}, \mathbf{x}_{\mathrm{rx}})$; $w, h$: Image resolution; $\alpha$: confidence regularisation coefficient}
\Output{Optimized parameters $\mu, \Sigma, \sigma, \vartheta, \phi$}

$\mu \leftarrow$ UniformInit(), $\Sigma, \sigma, \vartheta \leftarrow$ InitAttributes()
\hfill $\triangleright$ Initialize Gaussians and MLPs

$\phi \leftarrow$ InitConfidenceMLP()
\hfill $\triangleright$ Initialize confidence MLP $g_\phi: \mathbb{R}^3 \to \mathbb{R}$

\For{$i \leftarrow 1$ \KwTo $T$}{

  $(\mathbf{z}, \mathbf{x}_{\mathrm{rx}}) \leftarrow$ Sample$(D)$
  \hfill $\triangleright$ Load ground-truth spectrum and receiver position

  CullGaussians$(\mu, V)$
  \hfill $\triangleright$ Discard out-of-view Gaussians

  $\mu', \Sigma' \leftarrow$ TransformToView$(\mu, \Sigma, V)$
  \hfill $\triangleright$ Apply viewing transformation

  $G \leftarrow$ ProjectGaussians$(\mu', \Sigma')$
  \hfill $\triangleright$ Project Gaussians into 2D image plane

  $\hat{\mathbf{z}} \leftarrow 0$
  \hfill $\triangleright$ Initialize rendered spectrum

  \ForEach{pixel $(u, v)$}{

    $I \leftarrow$ IntersectGaussians$(G, u, v)$
    \hfill $\triangleright$ Find Gaussians overlapping pixel

    Sort $I$ by depth to receiver
    \hfill $\triangleright$ Back-to-front ordering

    $T \leftarrow 1$; \quad
    \ForEach{Gaussian $i$ in $I$}{
      $s_i \leftarrow f_{\vartheta}(\mathbf{x}_{\mathrm{rx}})$; \quad
      $\alpha_i \leftarrow \sigma_i \cdot \text{Vis}(G_i, u, v)$; \quad
      $\hat{\mathbf{z}}[u,v] \mathrel{+}= T \cdot \alpha_i \cdot s_i$; \quad
      $T \mathrel{*}= (1 - \alpha_i)$
    }
  }

  $\mathcal{L}_{\mathrm{task}} \leftarrow \mathcal{L}_{\mathrm{sp}}(\hat{\mathbf{z}},\, \mathbf{z})$
  \hfill $\triangleright$ Spectrum reconstruction loss (L1 + SSIM)

  $C \leftarrow 1 + \exp\!\left(g_\phi(\mathbf{x}_{\mathrm{rx}})\right)$
  \hfill $\triangleright$ Location-dependent confidence $C > 1$

  $\mathcal{L} \leftarrow C \cdot \mathcal{L}_{\mathrm{task}} - \alpha \log C$
  \hfill $\triangleright$ Confidence-weighted loss (Eq.~\ref{eq:conf_loss})

  $\nabla \mathcal{L} \leftarrow$ Backprop$(\mathcal{L})$
  \hfill $\triangleright$ Compute gradients

  $(\mu, \Sigma, \sigma, \vartheta, \phi) \leftarrow$ Adam$(\nabla \mathcal{L})$
  \hfill $\triangleright$ Update all parameters jointly

  \If{$i < T_{\mathrm{densify}}$}{

    AccumulateGradStats$(\mu, \nabla\mu)$
    \hfill $\triangleright$ Track per-Gaussian positional gradient norms

    \If{$i \bmod \Delta_{\mathrm{densify}} = 0$}{

      \ForEach{Gaussian $i$ with $\|\nabla\mu_i\| \geq \tau_{\mathrm{grad}}$}{

        \eIf{$\max(\Sigma_i) \leq \epsilon \cdot r_{\mathrm{scene}}$}{
          Clone Gaussian $i$
          \hfill $\triangleright$ Small Gaussian: duplicate in place
        }{
          Split Gaussian $i$ into $N$ children, scale $\leftarrow \Sigma_i / (0.8N)$
          \hfill $\triangleright$ Large Gaussian: replace with smaller children
        }
      }

      Prune Gaussians with $\sigma_i < \sigma_{\min}$ or screen radius $> r_{\max}$
      \hfill $\triangleright$ Remove transparent or oversized primitives

    }

    \If{$i \bmod \Delta_{\mathrm{reset}} = 0$}{
      Reset $\sigma_i \leftarrow \min(\sigma_i,\, 0.01)$ for all $i$
      \hfill $\triangleright$ Periodic opacity reset to remove floaters
    }
  }
}

\Return{$\mu, \Sigma, \sigma, \vartheta, \phi$}

\end{algorithm}

\paragraph*{Optimization.}
All Gaussian parameters $(\mu_i, \Sigma_i, \sigma_i)$ and the MLP weights $\vartheta$ are optimized
end-to-end using stochastic gradient descent. GSpaRC supports two reconstruction objectives
depending on the target output.

\textit{Channel reconstruction.}
For each training sample, GSpaRC renders the directional spectrum $\hat{\mathbf{z}}$ and obtains
the complex channel estimate $\hat{h}$ as its spatial mean (Eq.~\ref{eq:spatial_mean}).
The task loss is the complex MSE over the real and imaginary parts:
\begin{equation}
    \mathcal{L}_{\mathrm{ch}}
    =
    (h_{\mathrm{re}} - \hat{h}_{\mathrm{re}})^2 + (h_{\mathrm{im}} - \hat{h}_{\mathrm{im}})^2,
    \label{eq:loss_channel}
\end{equation}
which penalizes amplitude and phase discrepancies in the predicted channel.
Gradients flow through the spatial mean operator, the complex-valued volumetric compositing
rule, and the directional splatting pipeline.

\textit{Magnitude spectrum reconstruction.}
When the learning target is the magnitude spectrum $\mathbf{z}$ directly (e.g.\ RFID or BLE
measurements where phase is unavailable), the task loss combines pixel-wise fidelity and
perceptual structure:
\begin{equation}
    \mathcal{L}_{\mathrm{sp}}
    =
    (1 - \lambda_1)\,\mathcal{L}_{1}(\hat{\mathbf{z}}, \mathbf{z})
    +
    \lambda_1\,\bigl(1 - \mathrm{SSIM}(\hat{\mathbf{z}}, \mathbf{z})\bigr),
    \label{eq:loss_spectrum}
\end{equation}
where $\mathcal{L}_{1}$ is the mean absolute error, $\mathrm{SSIM}$ is the structural similarity
index, and $\lambda_1 = 0.2$ weights the two terms.
This formulation minimizes numerical error while preserving the directional structure of the
spatial spectrum, which is important for downstream beamforming and localization tasks.

\paragraph*{Densification, Pruning, and Implementation.}
To adapt the Gaussian set to the multipath structure of the scene, GSpaRC applies adaptive
densification during training. Gaussians whose positional gradient norm exceeds $\tau_{\mathrm{grad}}$
are under-reconstructed: small ones ($\max(\Sigma_i) \leq \epsilon \cdot r_{\mathrm{scene}}$) are
\emph{cloned} in place to fill gaps, while large ones are \emph{split} into $N$ smaller children
(scaled by $\Sigma_i / 0.8N$) to resolve fine-grained multipath features. Gaussians with opacity
below $\sigma_{\min}$ or footprint exceeding $r_{\max}$ are pruned, and periodic opacity resets
($\sigma_i \leftarrow \min(\sigma_i, 0.01)$) prevent saturation throughout training.
Gradients flow through a custom CUDA rasterization kernel that depth-sorts Gaussians and
accumulates complex contributions per direction pixel in parallel. Each Gaussian carries a dedicated
emission MLP ($\mathbb{R}^3\!\to\!\mathbb{R}^2$, 16 hidden units) stored in a flat per-Gaussian
buffer for fully batched GPU evaluation. All experiments use a single NVIDIA GeForce RTX~2080~Ti.

\paragraph*{Hyperparameters.}
Table~\ref{tab:optimizer_hyperparams} summarizes the key training hyperparameters. Position
learning rates are exponentially annealed; separate rates are used for opacity, scaling, rotation,
and the emission MLP to allow fine-grained convergence control.

\begin{table}[H]
\centering
\small
\setlength{\tabcolsep}{4pt}
\caption{GSpaRC training hyperparameters.}
\label{tab:optimizer_hyperparams}
\begin{tabular}{lc}
\toprule
\textbf{Parameter} & \textbf{Value} \\
\midrule
$\lambda_1$ (L1/SSIM weight)       & 0.2 \\
position\_lr (init $\to$ final)    & $1.6\!\times\!10^{-4} \to 1.6\!\times\!10^{-7}$ \\
opacity\_lr                        & 0.0055 \\
scaling\_lr                        & 0.005 \\
rotation\_lr                       & 0.001 \\
emission MLP lr                    & 0.005 \\
emission MLP architecture          & $3 \to 16 \to 2$ \\
\midrule
\multicolumn{2}{l}{\textit{Confidence MLP}} \\
$\alpha$                           & 0.3 \\
confidence\_lr                     & 0.005 \\
confidence architecture            & $3 \to 32 \to 1$ \\
\bottomrule
\end{tabular}
\end{table}

\paragraph*{Confidence-Weighted Optimization.}
Prediction difficulty varies across locations: positions near reflective surfaces or in
cluttered multipath environments are harder to reconstruct than line-of-sight locations. To capture this spatial heterogeneity, we augment either task loss with a
learned confidence score inspired by DUSt3R~\cite{wang2024dust3r}.
A compact MLP $g_\phi\colon \mathbb{R}^3 \to \mathbb{R}$ maps the receiver position
$\mathbf{x}_{\mathrm{rx}}$ to a scalar confidence
$C = 1 + \exp\!\left(g_\phi(\mathbf{x}_{\mathrm{rx}})\right)$,
which ensures $C > 1$. The floor at 1 prevents the network from fully suppressing any
training sample — even hard locations must contribute to the loss — following the
DUSt3R convention~\cite{wang2024dust3r}.
Denoting the active task loss as $\mathcal{L}_{\mathrm{task}} \in
\{\mathcal{L}_{\mathrm{ch}},\, \mathcal{L}_{\mathrm{sp}}\}$,
the final training objective is
\begin{equation}
    \mathcal{L}_{\mathrm{conf}}
    = C(\mathbf{x}_{\mathrm{rx}})\,\mathcal{L}_{\mathrm{task}}
      - \alpha\,\log C(\mathbf{x}_{\mathrm{rx}}),
    \label{eq:conf_loss}
\end{equation}
where $\alpha > 0$ is a regularization coefficient. Setting
$\partial\mathcal{L}_{\mathrm{conf}}/\partial C = 0$ gives $C^* = \alpha / \mathcal{L}_{\mathrm{task}}$
for unconstrained $C$: easy samples (low task loss) receive high confidence and are
up-weighted, while hard samples receive lower confidence. With the $C > 1$ constraint,
this optimum is attained when $\mathcal{L}_{\mathrm{task}} < \alpha$; for harder samples
where $\mathcal{L}_{\mathrm{task}} \geq \alpha$, the confidence saturates at 1.
We set $\alpha = 0.3$, which is above the 95th-percentile of the task loss, so the
unconstrained optimum is active for the vast majority of training samples.
The MLP $g_\phi$ uses a single hidden layer of 32 units and is
optimized jointly with all Gaussian parameters.
For reporting and threshold comparisons, the raw confidence $C > 1$ is
min-max normalized over the evaluation set: $\tilde{C} = (C - C_{\min})/(C_{\max} - C_{\min}) \in [0,1]$,
where $C_{\min}$ and $C_{\max}$ are the minimum and maximum values of $C$ across all test positions.
All figures and thresholds (e.g.\ $\tau = 0.59$) refer to $\tilde{C}$.

\section{Experimental Evaluation}
\label{sec:experiments}


We evaluate GSpaRC on three datasets: an indoor scene simulated with Sionna ray tracing for spectrum reconstruction under a fixed-TX, multiple-RX setting; a real-world RFID dataset for spectrum reconstruction~\cite{zhao2023nerf2}; and the real-world Argos massive-MIMO dataset for CSI reconstruction~\cite{shepard2016understanding}. Where applicable, we compare against NeRF\textsuperscript{2}~\cite{zhao2023nerf2}, WRFGS+~\cite{wen2025neuralrepresentationwirelessradiation}, and GSRF~\cite{yang2025gsrf}. All of these are geometry-aware representations, which have been empirically shown in \cite{yang2025gsrf} to outperform pure time-series approaches such as FIRE~\cite{liu2021fire}.


\subsection{Spectrum Reconstruction --- Sionna Simulation}
\label{subsec:sionna}

To evaluate GSpaRC in a controlled yet realistic propagation environment, we generate an indoor dataset using the Sionna ray tracer. The transmitter is fixed and multiple receiver positions are sampled throughout the room, a setting that aligns well with GSpaRC since the Gaussian scene representation is anchored to a persistent source. Each sampled TX--RX configuration yields a spectrum image at the receiver.

\begin{figure}[t]
\centering
\begin{subfigure}[t]{0.39\textwidth}
  \centering
  \includegraphics[width=\linewidth]{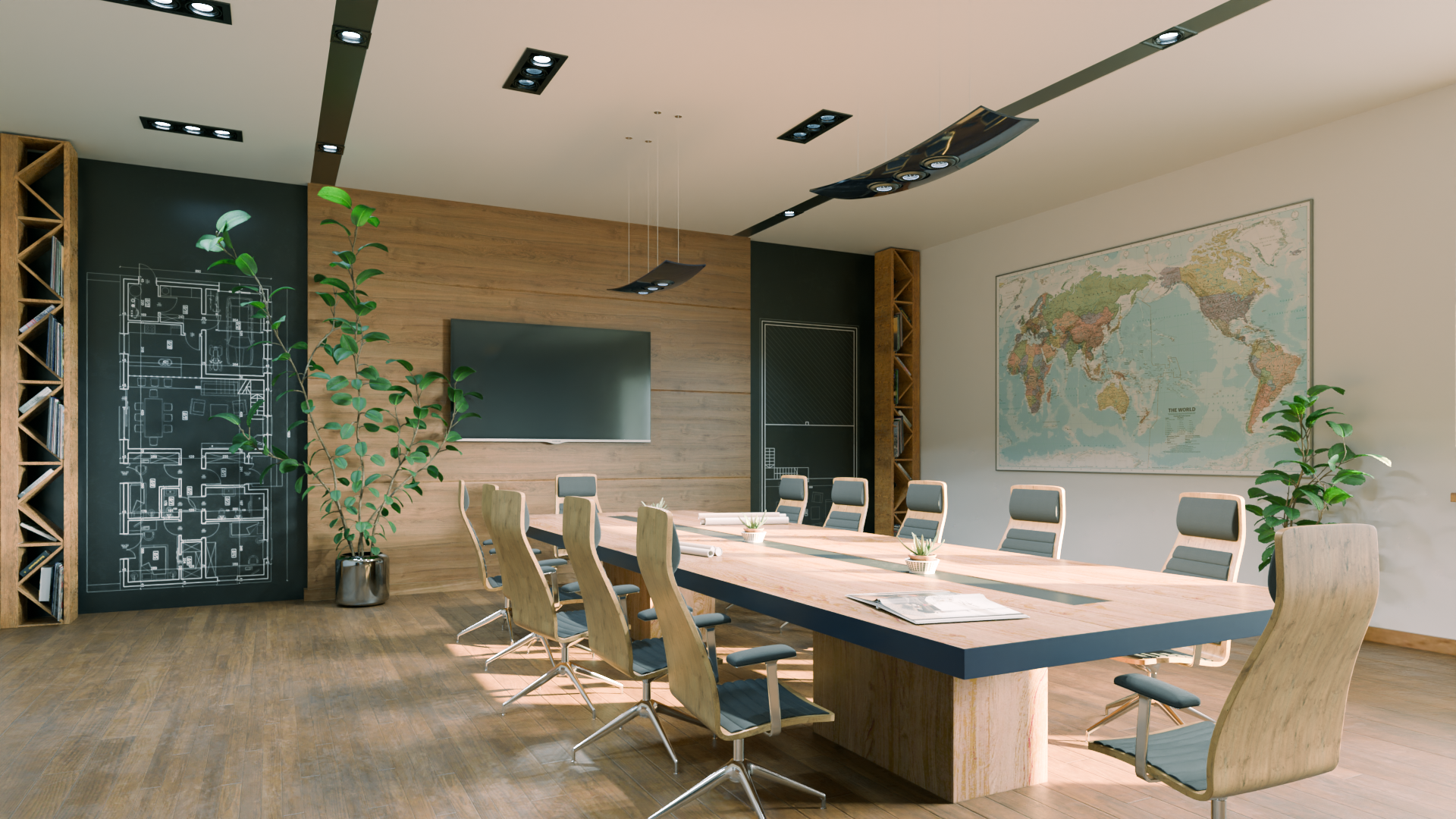}
  \caption{Conference room scene. The room has many objects and surfaces that create rich multi-path propagation.}
  \label{fig:sionna_scenes}
\end{subfigure}\hfill
\begin{subfigure}[t]{0.56\textwidth}
  \centering
  \includegraphics[width=\linewidth]{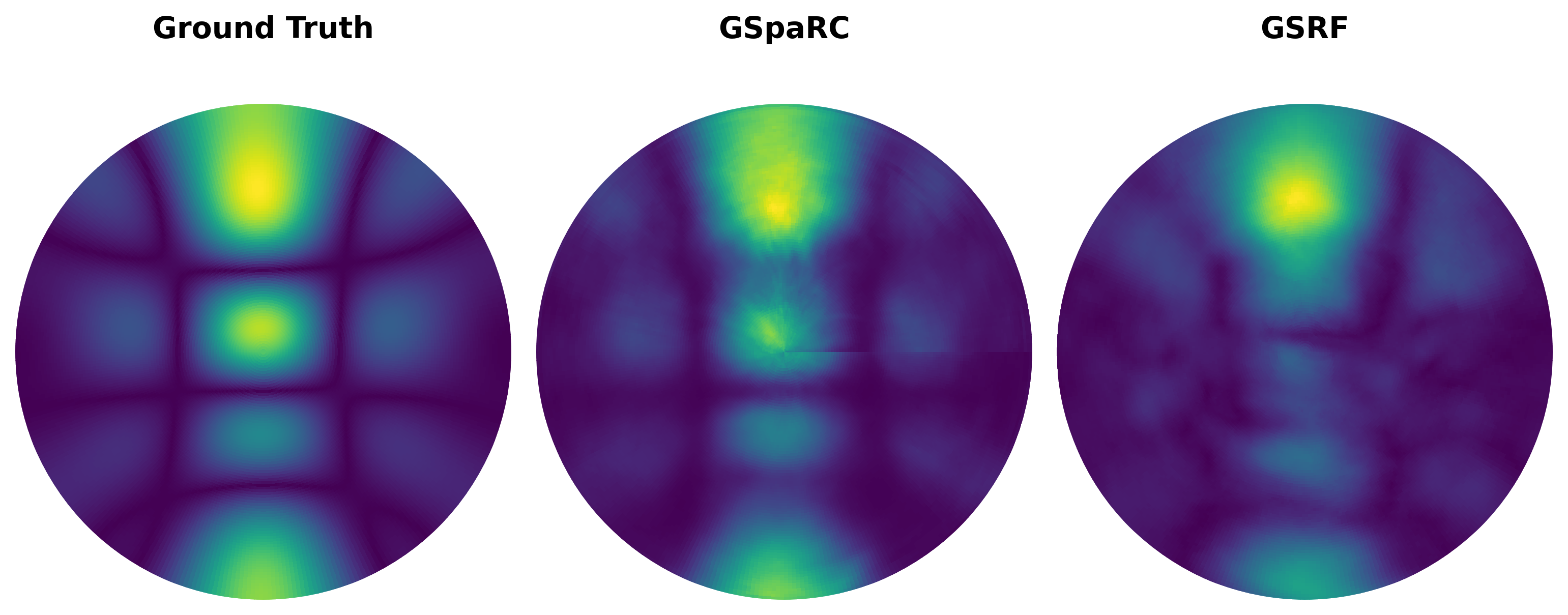}
  \caption{Spectrum reconstruction. Left to right: ground truth, GSpaRC, and GSRF for the same receiver position.}
  \label{fig:spectrum_comparison}
\end{subfigure}
\caption{Sionna conference room scene and representative spectrum reconstruction.}
\end{figure}

\smallskip\noindent\textbf{Conference Room Scene.}
This scene represents an indoor conference room spanning 14\,m~$\times$~10\,m with a 4\,m ceiling. The environment includes tables, chairs, glass partitions, and irregular wall materials that create rich multipath propagation. The transmitter is fixed near the ceiling of the room. The dataset contains 5,142 receiver positions uniformly sampled over the floor area, of which 1,543 are held out for testing. A visualization of this environment is shown in Fig.~\ref{fig:sionna_scenes}.

\begin{figure}[H]
\centering
\begin{subfigure}[t]{0.41\textwidth}
  \centering
  \includegraphics[width=\linewidth]{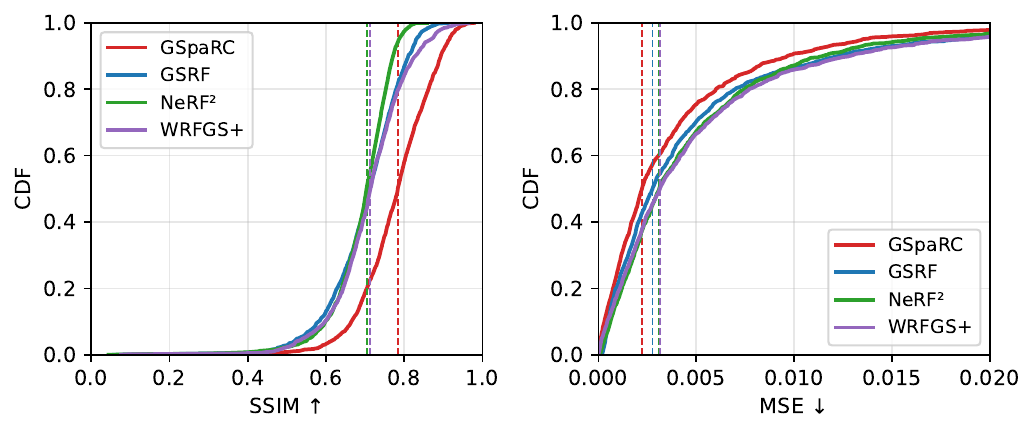}
  \caption{CDF of SSIM and MSE (1,543 test positions). GSpaRC outperforms all baselines. Dashed lines: medians.}
  \label{fig:sionna_cdf}
\end{subfigure}\hfill
\begin{subfigure}[t]{0.54\textwidth}
  \centering
  \includegraphics[width=\linewidth]{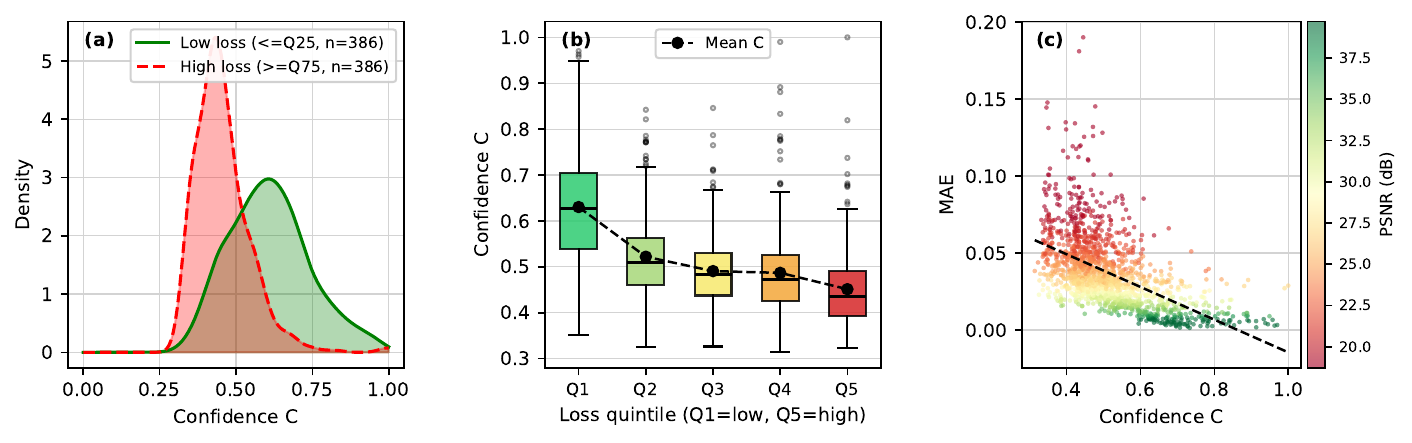}
  \caption{Confidence--reconstruction relationship. (a) Low/high-loss distributions separated. (b) Confidence vs.\ loss quintiles. (c) Confidence vs.\ MAE.}
  \label{fig:scene1_confidence}
\end{subfigure}
\caption{Sionna conference room: performance CDF and confidence calibration.}
\end{figure}

Table~\ref{tab:algo_comparison_sionna} presents a comparison of GSpaRC with all baseline methods for this scene. GSpaRC consistently achieves the best overall performance across all evaluated metrics, demonstrating superior reconstruction quality relative to prior approaches. Figure~\ref{fig:sionna_cdf} further highlights these improvements through the CDFs of SSIM and MSE, indicating that the gains from GSpaRC are consistent across the full distribution of test locations. Finally, Figure~\ref{fig:scene1_confidence} demonstrates that GSpaRC’s confidence estimates are well aligned with reconstruction quality.


\begin{table}[H]
\centering
\caption{Quantitative comparison of algorithms on the Sionna conference room dataset.}
\begin{tabular}{@{}lccccc@{}}
\toprule
\textbf{Algorithm} & \multicolumn{2}{c}{\textbf{SSIM} $\uparrow$} &
                     \multicolumn{2}{c}{\textbf{MSE} $\downarrow$} &
                     \textbf{Train} $\downarrow$ \\
\cmidrule(lr){2-3}\cmidrule(lr){4-5}
 & Mean & Median & Mean & Median & (hrs)\\
\midrule
NeRF\textsuperscript{2}  & 0.69 & 0.70 & 0.00513 & 0.00310 & 4.15 \\
WRFGS+                  & 0.71 & 0.71 & 0.00550 & 0.00314 & 3.67 \\
GSRF                    & 0.70 & 0.71 & 0.00512 & 0.00276 & 1.1 \\
\hline
\textbf{GSpaRC} & \textbf{0.78} & \textbf{0.78} & \textbf{0.00417} & \textbf{0.00223} & \textbf{1} \\
\bottomrule
\end{tabular}
\label{tab:algo_comparison_sionna}
\end{table}


\subsection{Spectrum Reconstruction --- RFID Dataset}
\label{subsec:rfid}

We evaluate GSpaRC on the publicly available RFID Spectrum NeRF\textsuperscript{2} dataset~\cite{zhao2023nerf2}. The receiver is fixed, operates at 915~MHz, and is equipped with a $4{\times}4$ antenna array. The transmitter, an RFID tag, is placed at 6,123 distinct locations, and each location yields a spatial spectrum image recorded at the receiver. The loss is a weighted combination of L1 and SSIM with $\lambda_1 = 0.2$.

Table~\ref{tab:algo_comparison_full} presents a quantitative comparison using SSIM and MSE. GSpaRC matches GSRF in mean SSIM (0.82) and achieves the fastest rendering time at 0.8~ms, more than 20$\times$ faster than GSRF (23\,ms), and the lowest latency among all methods. GSRF attains better MSE and PSNR when trained to convergence over 27~minutes, while GSpaRC converges in a comparable time with substantially lower inference cost.

\begin{table}[H]
\centering
\small
\caption{Quantitative and efficiency comparison on the RFID dataset.}
\begin{tabular}{@{}lccccc@{}}
\toprule
\textbf{Algorithm} 
& \textbf{SSIM} $\uparrow$ 
& \textbf{MSE} $\downarrow$ 
& \textbf{PSNR (dB)} $\uparrow$ 
& \textbf{Train} 
& \textbf{Render} \\

& Mean 
& Mean 
& Mean 
& (hrs) $\downarrow$ 
& (ms) $\downarrow$ \\
\midrule

WRFGS+ & 0.78 & 0.0112 & {21.14} & 2.0 & {8} \\
NeRF\textsuperscript{2} & 0.75 & {0.0101} & 21.09 & 8+ & 230 \\
GSRF & {0.82} & {0.0094} & {22.05} & {0.45} & 23 \\
\midrule
GSpaRC & \textbf{0.82} & \textbf{0.0134} & \textbf{20.35} & \textbf{0.5} & \textbf{0.8} \\

\bottomrule
\end{tabular}
\label{tab:algo_comparison_full}
\end{table}
\subsection{CSI Reconstruction --- Argos Dataset}
\label{subsec:argos}

Beyond spectrum reconstruction, GSpaRC can predict complex-valued CSI at unseen positions. We evaluate this ability on the Argos massive-MIMO dataset~\cite{shepard2016understanding}, a real-world outdoor deployment with a 64-antenna base station at 2.4~GHz and 4,000 user positions. Following~\cite{liu2021fire}, we use a pretrained CSI encoder to map uplink measurements to 3D positions (since the receiver positions are not explicitly provided in the dataset), with the massive-MIMO base-station antenna serving as the fixed transmitter. GSpaRC is trained to predict the complex downlink CSI, both real and imaginary, at each position over 26 subcarriers, using a complex MSE loss with a learned uncertainty term for per-position confidence.

The dataset is split into 3,200 training and 800 test samples. We report SNR~(dB) and NMSE $= \sum\|\hat{h}-h\|^2 / \sum\|h\|^2$, evaluated under the same normalization as GSRF.   Table~\ref{tab:argos_csi} shows our results indicating that the accuracy of GSpaRC is similar to that of GSRF, while significantly decreasing the inference time. 

\begin{table}[H]
\centering
\caption{CSI reconstruction on the Argos dataset (antenna 0, 800 test samples).}
\begin{tabular}{@{}lcccc@{}}
\toprule
& \multicolumn{2}{c}{\textbf{SNR (dB)} $\uparrow$} & \textbf{NMSE} $\downarrow$ & \textbf{Render} $\downarrow$ \\
\cmidrule(lr){2-3}
\textbf{Method} & \textbf{1-SC} & \textbf{26-SCs} &  &  \textbf{(ms)}\\
\midrule
GSRF   & 23.62          & 20.87          & 0.0247          & 22.76 \\
\textbf{GSpaRC} & \textbf{23.52} & \textbf{19.59} & \textbf{0.0110} & \textbf{3.86} \\
\bottomrule
\end{tabular}
\label{tab:argos_csi}
\end{table}

\begin{figure}[t]
\centering
\includegraphics[width=0.75\textwidth]{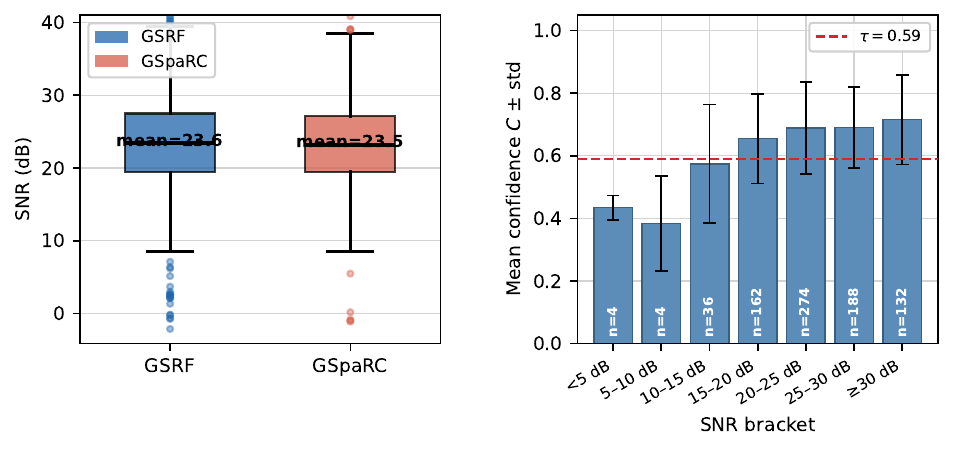}
\caption{Left: SNR distribution over 800 test positions (Argos, antenna~0, subcarrier~0). GSpaRC achieves a mean SNR comparable to that of GSRF (23.5~dB vs.\ 23.6~dB). Right: mean GSpaRC confidence by SNR bracket. Confidence increases monotonically with SNR, and the $\tau=0.59$ threshold (dashed) separates low-SNR from high-SNR samples.}
\label{fig:csi_boxplot}
\end{figure}

In addition to CSI values, GSpaRC outputs a per-position \emph{confidence} score learned via an uncertainty-aware objective: positions with small prediction error receive high confidence, while those with large error receive low confidence. As shown in Figure~\ref{fig:confidence_coverage}, at threshold $\tau=0.59$, approximately 70\% of the 800 test positions are assigned high confidence, meaning that pilot transmissions can be eliminated at those locations. This calibrated uncertainty estimate is used directly in the downstream tasks of Section~\ref{sec:downstream}.
\section{Downstream Applications}
\label{sec:downstream}

Beyond CSI reconstruction accuracy, an important question is whether the CSI predicted by GSpaRC is directly useful for communication tasks. We evaluate two downstream PHY-layer applications: (i) link adaptation through modulation and coding scheme (MCS) selection and (ii) end-to-end physical-layer communication, using the real-world Argos dataset within the Sionna simulator. In both cases, GSpaRC outputs the predicted complex CSI $(\hat{h}_{\mathrm{re}},\, \hat{h}_{\mathrm{im}})$ and a confidence score $C$ for each receiver location. This confidence estimate is critical because prediction quality is not uniform across space: locations with sparse training support or more complex propagation conditions are harder to predict reliably. We therefore use confidence-aware filtering, in which only samples with $C \ge \tau$ use predicted CSI, while the remaining low-confidence samples fall back to conventional pilot-based estimation.

\subsection{Confidence-Aware Link Adaptation}
\label{subsec:mcs}

We first evaluate whether GSpaRC-predicted CSI is accurate enough for link adaptation. In practical systems, the modulation and coding scheme (MCS) is selected based on channel quality, and inaccurate CSI can either cause decoding failure through overly aggressive MCS selection or reduce spectral efficiency through overly conservative choices. Our goal is to determine whether GSpaRC can support this decision while using its confidence score to identify the receiver locations where prediction quality is reliable.

For this experiment, we use Sionna's link adaptation framework. For each receiver location, the predicted CSI is used to compute an effective SINR, which is then mapped to an MCS subject to a target BLER of 0.1. To keep the operating point meaningful for link adaptation, noise is added so that the effective SINR spans a practical range. The selected MCS is then compared against the oracle MCS obtained from the corresponding ground-truth CSI.

\begin{figure}[H]
\centering
\begin{subfigure}[t]{0.49\textwidth}
  \centering
  \includegraphics[width=\linewidth]{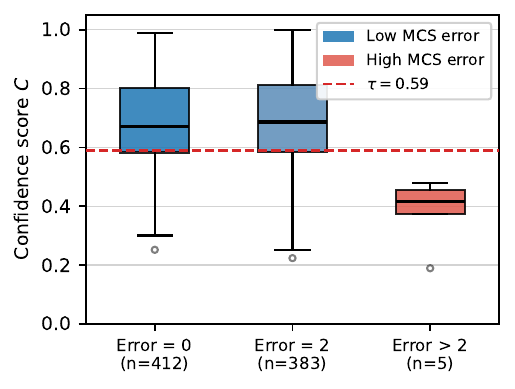}
  \label{fig:mcs}
\end{subfigure}\hfill
\begin{subfigure}[t]{0.49\textwidth}
  \centering
  \includegraphics[width=\linewidth]{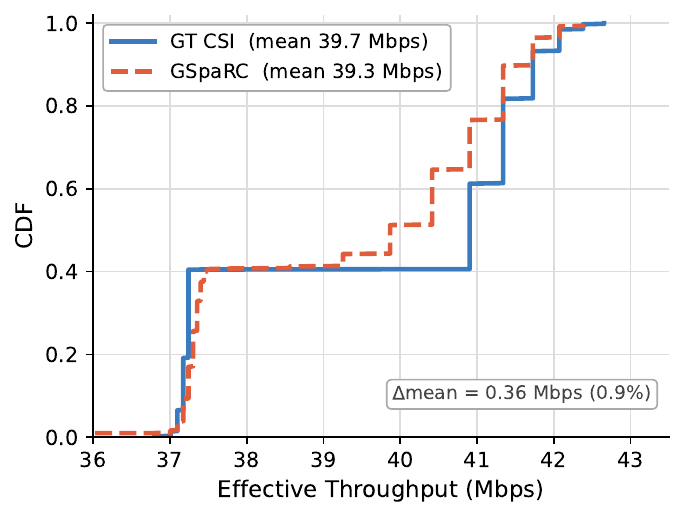}
  \label{fig:mcs_throughput}
\end{subfigure}
\caption{(a) Confidence distributions grouped by MCS prediction error. Low-error samples lie above $\tau=0.59$, while the rare high-error samples ($n{=}5$, error${>}2$) fall entirely below it. (b) GSpaRC-based MCS selection achieves a mean throughput of 39.3\,Mbps vs.\ 39.7\,Mbps for GT CSI.}
\label{fig:mcs_combined}
\end{figure}

Figure~\ref{fig:csi_boxplot} (right) shows that confidence increases monotonically with SNR, and the $\tau=0.59$ threshold cleanly separates the low-SNR brackets (below 15~dB) from the high-SNR majority, confirming that the score reflects genuine reconstruction quality. Figure~\ref{fig:mcs_combined}(a) groups test samples by MCS prediction error and shows that low-error samples (error~$=0$ and error~$=2$) lie above $\tau=0.59$, while the rare high-error samples (error~$>2$, only 5 samples) fall entirely below it. For each sample, the ideal throughput is computed from the assigned MCS as $T_{\text{ideal}} = (M \times R)\, N_{\text{sc}}\, f_s\, N_L$, where $M$ is the modulation order, $R$ the code rate, $N_{\text{sc}} = 624$, $f_s = 28{,}000$ symbols/s, and $N_L$ the number of layers. The effective throughput is then $T_{\text{eff}} = T_{\text{ideal}} (1 - \text{TBLER})$, where TBLER is estimated using Sionna's PHY abstraction from the per-sample SINR. Figure~\ref{fig:mcs_combined}(b) shows that GSpaRC-based MCS selection achieves a mean throughput of 39.3\,Mbps versus 39.7\,Mbps for GT CSI, a gap of only 0.36\,Mbps (0.9\%), confirming near-optimal link throughput.

\subsection{Confidence-Aware Communication}
\label{subsec:ber}

We next evaluate whether GSpaRC-predicted CSI can be used directly in a practical communication pipeline. To do so, we use test samples from the Argos CSI dataset and feed the predicted CSI into Sionna's end-to-end 5G NR PUSCH simulation. The receiver processing chain includes DMRS-based channel estimation, LMMSE equalization, and decoding. Performance is evaluated in terms of both Bit Error Rate (BER) and Block Error Rate (BLER) as a function of $E_b/N_0$, the ratio of energy per information bit to noise power spectral density, which captures the effective signal-to-noise ratio. 

\begin{figure}[t]
\centering
\includegraphics[width=\columnwidth]{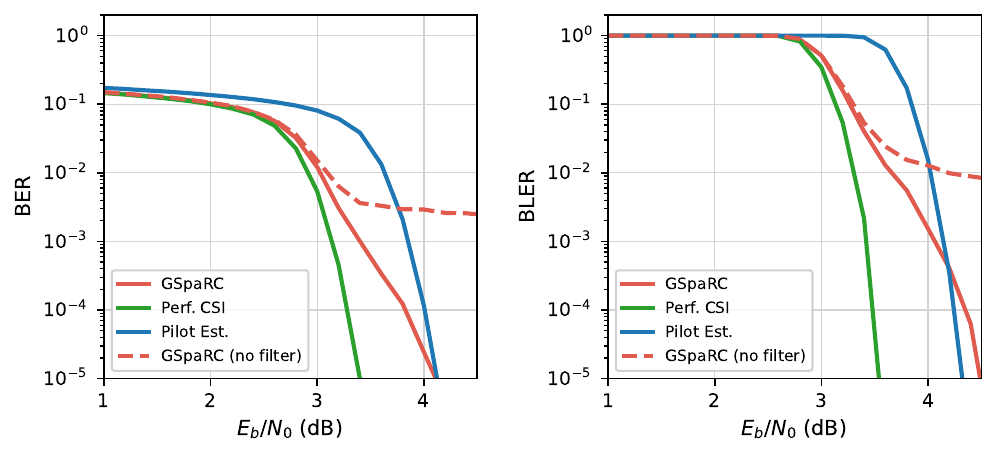}
\caption{BER (left) and BLER (right) versus $E_b/N_0$ in Sionna's end-to-end 5G NR PUSCH simulation. Without confidence filtering, GSpaRC-predicted CSI degrades noticeably. After applying the threshold $C \ge \tau=0.59$ (solid), the GSpaRC curves closely match perfect CSI, while pilot-based LS estimation requires a higher $E_b/N_0$ to attain the same error rate.}
\label{fig:ber_snr}
\end{figure}
Figure~\ref{fig:ber_snr} compares three receiver-side CSI cases: i) perfect CSI knowledge, ii) CSI estimated from DMRS pilots using a least-squares (LS) estimator, and iii) CSI predicted by GSpaRC. A key observation is that downstream link performance is strongly tied to the confidence score produced by GSpaRC. When all test samples are used without filtering, prediction errors at difficult receiver locations noticeably degrade communication performance. In contrast, after applying a confidence threshold of $\tau=0.59$, approximately 70\% of the Argos test samples are retained. For this high-confidence subset, the GSpaRC-based BER and BLER curves for QPSK closely follow the perfect-CSI and pilot-based LS baselines. This shows that GSpaRC can provide CSI of sufficient quality for PHY-layer communication over a substantial fraction of receiver locations, while low-confidence samples can be naturally identified for fallback to conventional pilot-based estimation.

\section{Ablation Studies}
\label{sec:ablation}

We isolate three design choices in GSpaRC: the physics-informed distance term, the CUDA rendering pipeline, and the capacity of the per-Gaussian emission MLP. These ablations show that the RF-specific components improve accuracy, while the learned decoder can remain small enough to preserve low-millisecond inference.

\subsection{Impact of Physics-Informed Distance Weighting}
\label{subsec:ablation-distance}

We first evaluate the contribution of the $1/d$ free-space attenuation term in the RF compositing rule. This term gives the renderer a simple physical prior: Gaussians farther from the transmitter should contribute less energy. We compare GSpaRC with and without this term on the RFID dataset.

\begin{figure}[H]
\centering
\captionsetup[subfigure]{labelfont=bf,font=small}
\begin{subfigure}[t]{0.32\textwidth}
    \centering
    \includegraphics[width=\linewidth]{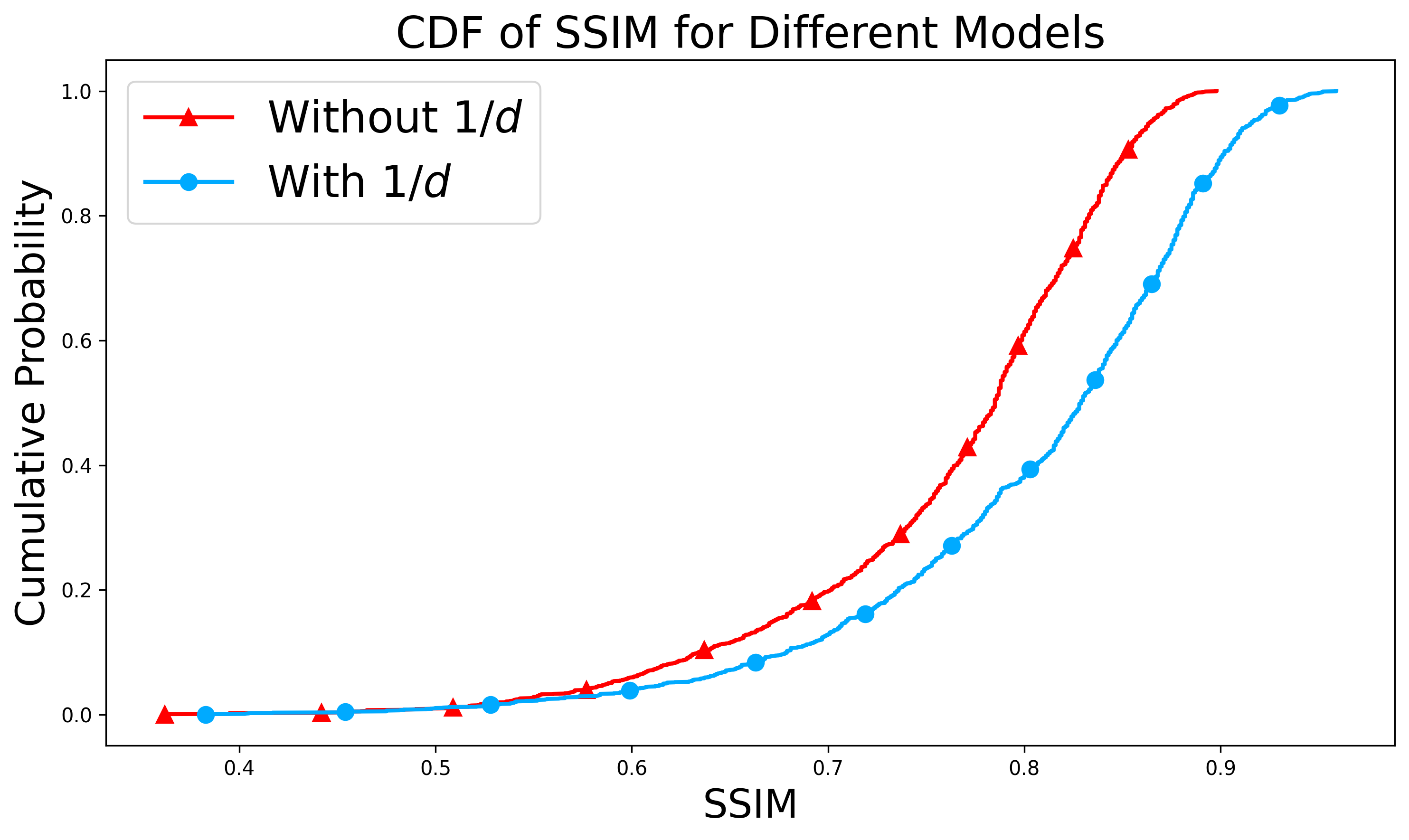}
    \caption{SSIM ($\uparrow$)}
\end{subfigure}
\hfill
\begin{subfigure}[t]{0.32\textwidth}
    \centering
    \includegraphics[width=\linewidth]{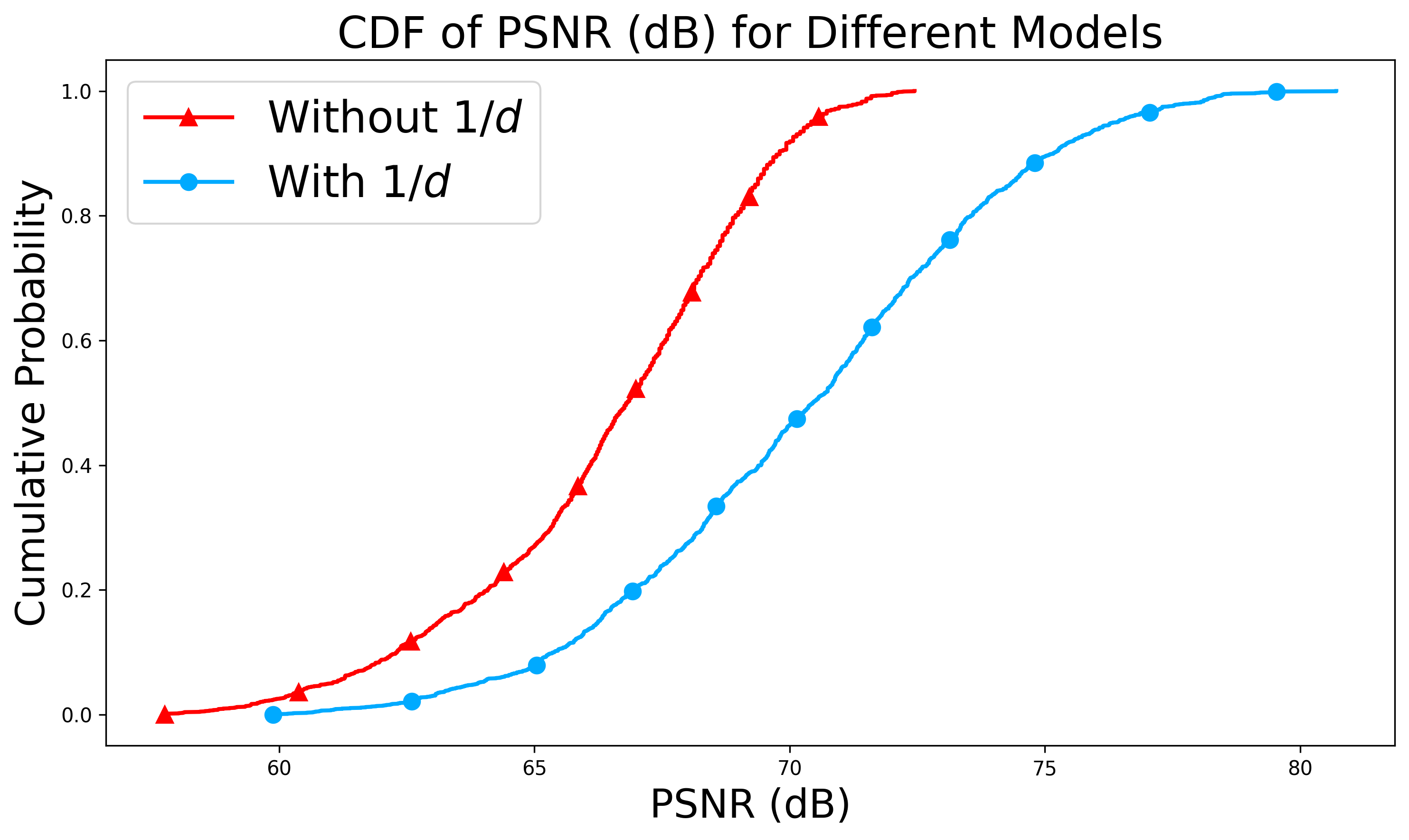}
    \caption{PSNR ($\uparrow$)}
\end{subfigure}
\hfill
\begin{subfigure}[t]{0.32\textwidth}
    \centering
    \includegraphics[width=\linewidth]{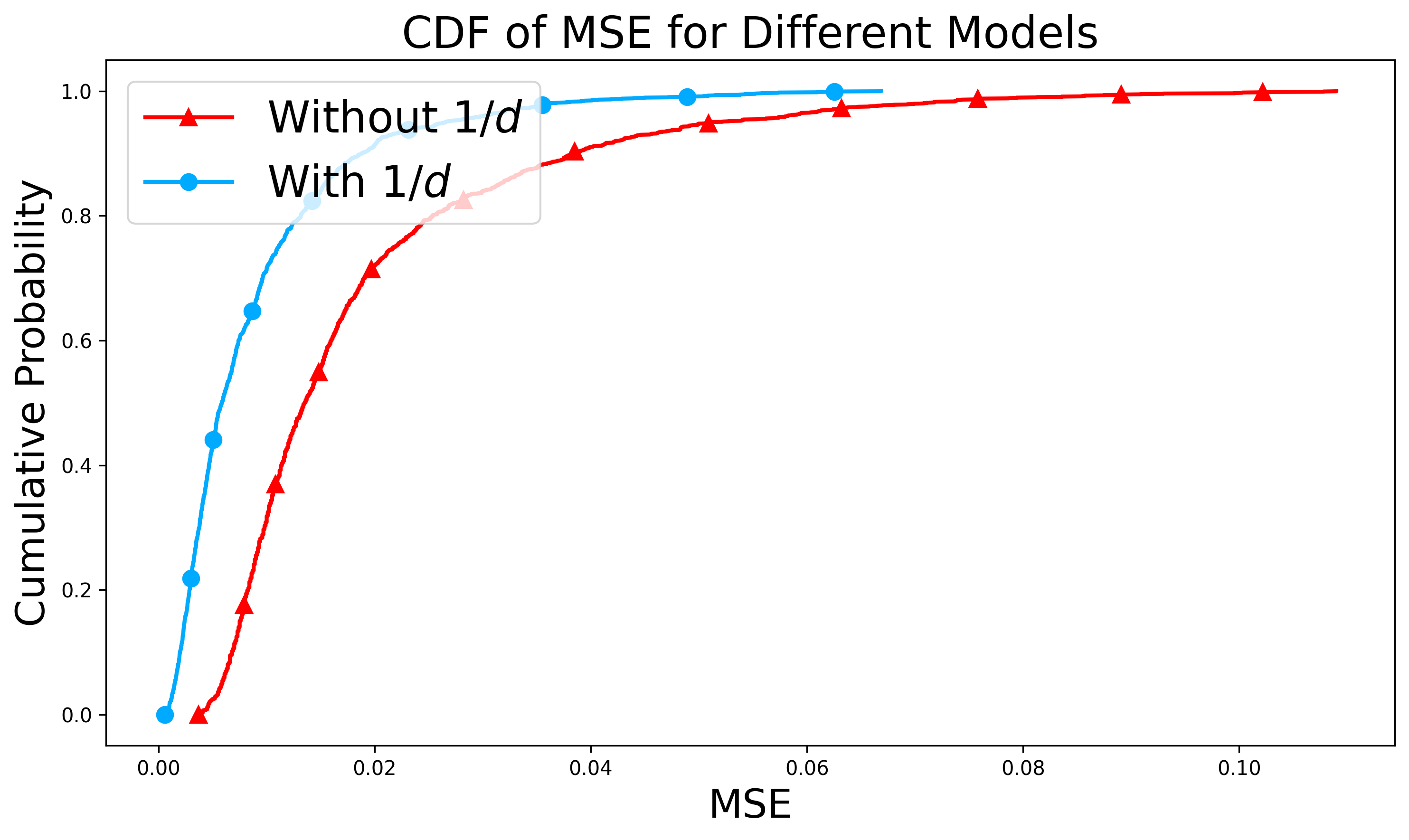}
    \caption{MSE ($\downarrow$)}
\end{subfigure}
\caption{Effect of physics-informed distance weighting on the RFID dataset. Adding the $1/d$ attenuation term improves SSIM and PSNR while reducing MSE, showing that a simple path-loss prior improves reconstruction quality.}
\label{fig:ssim_cdf_physics_informed}
\end{figure}

Figure~\ref{fig:ssim_cdf_physics_informed} shows that including $1/d$ shifts the SSIM and PSNR CDFs to the right and the MSE CDF to the left. Thus, the physics-informed model improves both perceptual and numerical reconstruction quality. The gain is most visible in the high-quality part of the distribution, where accurate modeling of attenuation helps sharpen the dominant spatial lobes. This result supports the main modeling choice in GSpaRC: simple RF priors can reduce the burden on the learned Gaussian representation and improve reconstruction without adding significant computation.

\subsection{Rendering Time Analysis}
\label{subsec:ablation-render}

\begin{figure}[t]
\centering
\includegraphics[width=\textwidth]{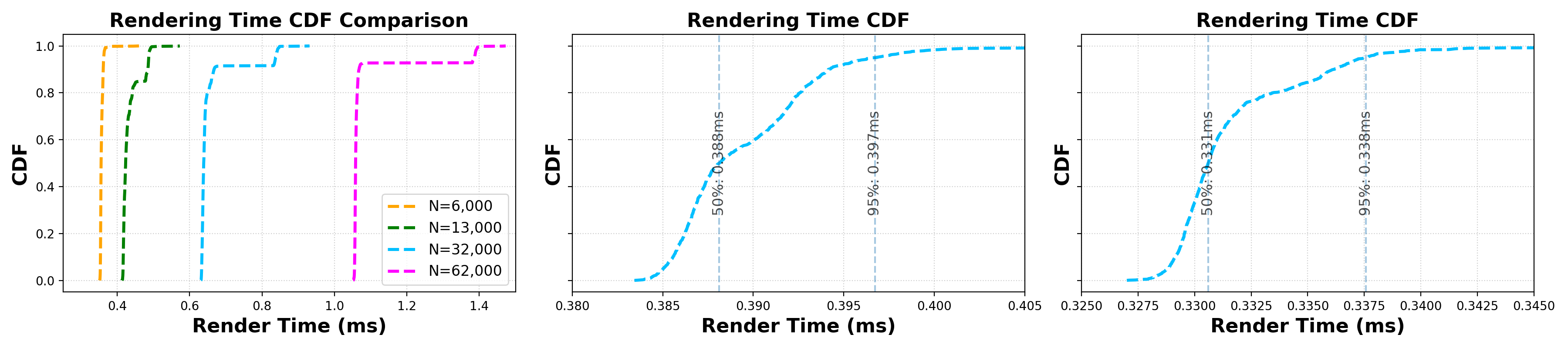}
\caption{Rendering-time CDFs. \textbf{Left:} Latency (in milliseconds) as the number of Gaussians increases. \textbf{Middle:} RFID dataset. \textbf{Right:} Sionna dataset. The custom CUDA renderer maintains low-millisecond latency across datasets and scales smoothly with representation size.}
\label{fig:cdf_combined}
\end{figure}

We next study whether the custom CUDA renderer preserves low latency as scene complexity changes. Figure~\ref{fig:cdf_combined} reports rendering-time CDFs for different Gaussian counts and for the RFID and Sionna datasets.\footnote{Figure~\ref{fig:cdf_combined} reports CUDA kernel time only; render times in Tables~\ref{tab:algo_comparison_sionna}--\ref{tab:argos_csi} include end-to-end inference with MLP evaluation and data transfer.} The left panel shows that latency increases smoothly with the number of Gaussians, rather than exhibiting sharp slowdowns. Even for the largest tested representation, with roughly $6.2\times 10^4$ Gaussians, rendering remains in the low-millisecond range.

The dataset-specific CDFs show the same trend. On RFID, the median and 95th-percentile rendering times are approximately 0.39\,ms and 0.40\,ms. On Sionna, they are approximately 0.33\,ms and 0.34\,ms. These results indicate that GSpaRC can render spectra well within the few-millisecond time scale needed for fast channel prediction and related PHY-layer decisions, even on a modest GPU.

\subsection{Effect of Emission MLP Size}
\label{subsec:ablation-mlp-size}

Finally, we vary the hidden size of the per-Gaussian emission MLP. This ablation tests whether GSpaRC needs a large neural decoder or whether most of the useful structure is already captured by the geometry of the Gaussian field.

\begin{figure}[t]
\centering
\includegraphics[width=\textwidth]{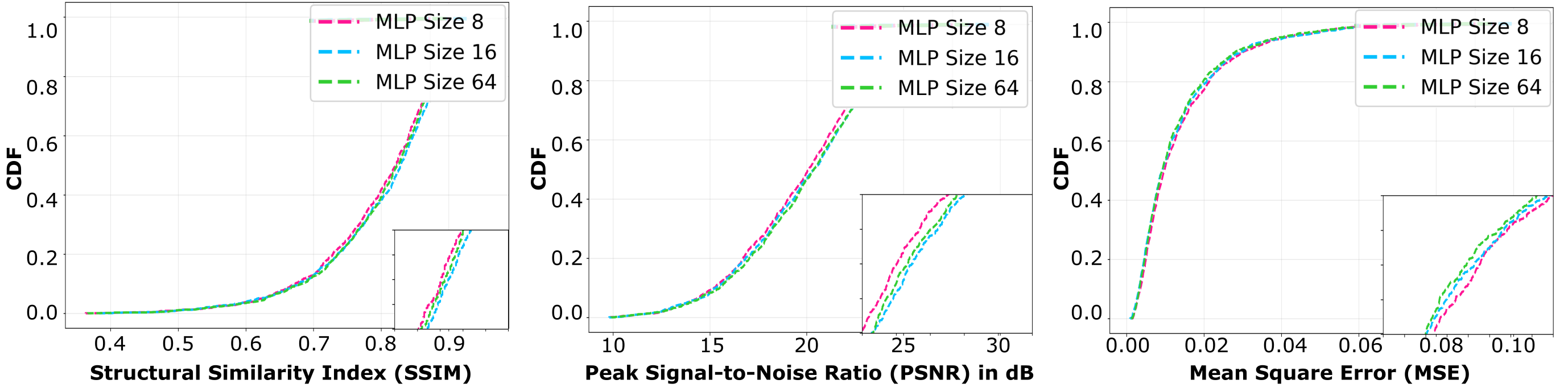}
\caption{Effect of per-Gaussian emission MLP size. The CDFs of SSIM, PSNR, and MSE are similar for hidden sizes 8, 16, and 64, indicating that a small decoder is sufficient once the geometry-aware Gaussian representation is learned.}
\label{fig:mlp_size}
\end{figure}

Figure~\ref{fig:mlp_size} compares hidden sizes 8, 16, and 64 using SSIM, PSNR, and MSE CDFs. The curves are nearly overlapping across all three metrics. Increasing the MLP size produces only minor changes in the high-accuracy tail and does not substantially improve the overall distribution. This shows that reconstruction quality is driven mainly by the spatial Gaussian representation and RF-aware rendering, rather than by a large neural network. We therefore use a hidden size of 16, which provides a good accuracy and efficiency tradeoff.

\section{Limitations}
While \textbf{GSpaRC} demonstrates promising performance for real-time RF channel reconstruction, several limitations remain. First, like other radiance field-based approaches, GSpaRC requires accurate receiver localization. This is needed to correctly orient the Gaussian primitives during rendering and directly affects the accuracy of the resulting spectrum prediction. Although localization is available in many practical settings, for example from IMUs or infrastructure-based tracking, errors in receiver position can degrade reconstruction fidelity. Second, the current GSpaRC formulation assumes a relatively static environment during training and inference. Rapid geometric changes, such as moving objects or people, may violate this assumption and reduce accuracy unless the model is retrained frequently or extended with temporal dynamics. Finally, although GSpaRC achieves low latency on modest hardware, it still relies on GPU acceleration and a pre-training phase that may not be practical for highly constrained edge devices. Addressing these limitations, especially under localization uncertainty and in dynamic environments, is an important direction for future work toward making GSpaRC more robust and more widely deployable.

\section{Conclusion}

We presented \textbf{GSpaRC}, a framework for real-time wireless channel reconstruction that combines the speed and differentiability of 3D Gaussian splatting with domain-specific insights from RF propagation. GSpaRC models the environment using a compact set of learned Gaussian primitives, augmented with physics-informed neural features, and employs a custom CUDA-based rendering pipeline optimized for hemispherical RF signal reception. This design enables efficient low-latency inference and supports high-fidelity spectrum reconstruction from sparse RF measurements.

Our evaluations show that GSpaRC achieves competitive accuracy across SSIM, PSNR, and MSE while reducing training and inference time by an order of magnitude relative to state-of-the-art methods such as NeRF2 and WRF-GS+. Importantly, GSpaRC is the first splatting-based approach to operate in the low-millisecond regime needed for practical wireless channel estimation, making it a scalable and realistic solution for modern wireless systems.

By aligning fast, differentiable 3D reconstruction techniques with the physical requirements of RF propagation and antenna reception, GSpaRC lays the groundwork for real-time, data-driven channel estimation in latency-sensitive wireless applications. Its potential extends to digital twins, adaptive beamforming, and mobility-aware communication in 5G and future networks.

\bibliography{bib/references}
\bibliographystyle{unsrt}

\end{document}